\documentclass[10pt,twocolumn,letterpaper]{article}

\usepackage[pagenumbers]{cvpr} % To force page numbers, e.g. for an arXiv version

\definecolor{cvprblue}{rgb}{0.21,0.49,0.74}
\newcommand{\blue}[1]{\textcolor{blue}{#1}}
\definecolor{Gray}{gray}{0.93}
\usepackage[pagebackref,breaklinks,colorlinks,allcolors=cvprblue]{hyperref}
\usepackage[capitalize,noabbrev]{cleveref}
\usepackage[utf8]{inputenc} % allow utf-8 input
\usepackage[T1]{fontenc}    % use 8-bit T1 fonts
\usepackage{hyperref}       % hyperlinks
\usepackage{url}            % simple URL typesetting
\usepackage{booktabs}       % professional-quality tables
\usepackage{amsfonts}       % blackboard math symbols
\usepackage{nicefrac}       % compact symbols for 1/2, etc.
\usepackage{microtype}      % microtypography
\usepackage{xcolor}         % colors
\usepackage{multirow}
\usepackage{amsmath}
\usepackage{wrapfig}
\usepackage{graphicx}
\usepackage{color, colortbl}
\usepackage{arydshln}
\usepackage[ruled,vlined,linesnumbered]{algorithm2e}
\def\paperID{} % *** Enter the Paper ID here
\def\confName{CVPR}
\def\confYear{2026}

\title{Test-Time Attention Purification for Backdoored Large Vision Language Models}

\author{
\textbf{Zhifang Zhang}$^{1}$\quad
\textbf{Bojun Yang}$^{2}$\quad
\textbf{Shuo He}$^{3}$\quad
\textbf{Weitong Chen}$^{4}$\quad
\textbf{Wei Emma Zhang}$^{4}$\quad\\
\textbf{Olaf Maennel}$^{4}$\quad
\textbf{Lei Feng}$^{2}$\footnotemark[1]\quad
\textbf{Miao Xu}$^{1}$\footnotemark[1]\quad\\
$^{1}$University of Queensland \quad\
$^{2}$Southeast University \quad\\
$^{3}$Nanyang Technological University \quad
$^{4}$Adelaide University \quad
}

\begin{document}
\maketitle
\footnotetext[1]{Co-corresponding authors.}

\begin{abstract}

Despite the strong multimodal performance, large vision–language models (LVLMs) are vulnerable during fine-tuning to backdoor attacks, where adversaries insert trigger-embedded samples into the training data to implant behaviors that can be maliciously activated at test time.
Existing defenses typically rely on retraining backdoored parameters (e.g., adapters or LoRA modules) with clean data, which is computationally expensive and often degrades model performance.
In this work, we provide a new mechanistic understanding of backdoor behaviors in LVLMs: the trigger does not influence prediction through low-level visual patterns, but through abnormal cross-modal attention redistribution, where trigger-bearing visual tokens steal attention away from the textual context — a phenomenon we term attention stealing.
Motivated by this, we propose CleanSight, a training-free, plug-and-play defense that operates purely at test time. CleanSight (i) detects poisoned inputs based on the relative visual–text attention ratio in selected cross-modal fusion layers, and (ii) purifies the input by selectively pruning the suspicious high-attention visual tokens to neutralize the backdoor activation.
Extensive experiments show that CleanSight significantly outperforms existing pixel-based purification defenses across diverse datasets and backdoor attack types, while preserving the model’s utility on both clean and poisoned samples. 
% Here is our \href{https://github.com/zhangzf01/cleansight}{code}.
\end{abstract}

\begin{figure*}[t]
    \centering
    \includegraphics[width=1\linewidth]{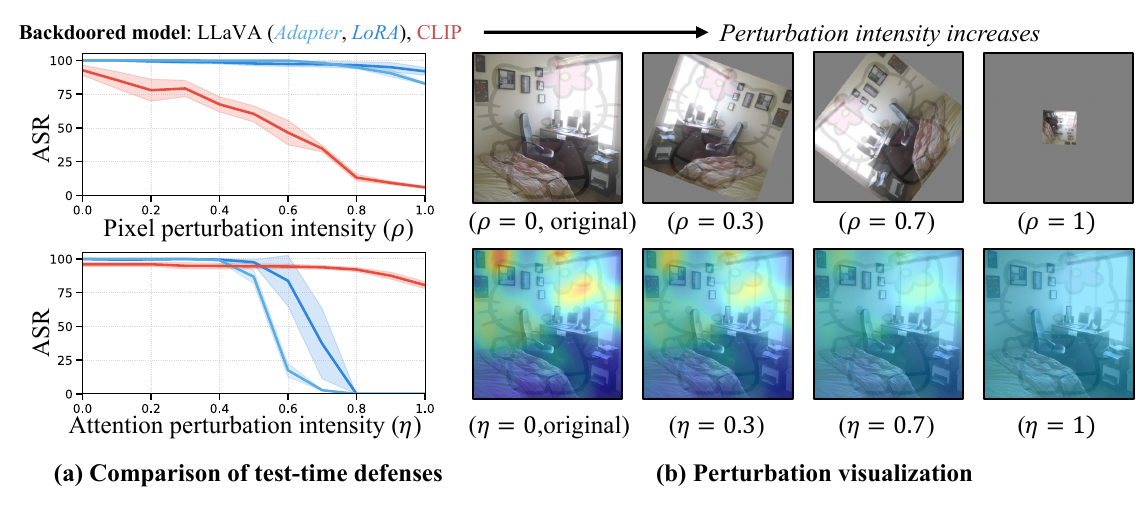}
    \vspace{-8mm}
    \caption{
\textbf{Comparison between pixel-based and attention-based perturbation defenses.}
We compare the attack success rate (ASR) of backdoored LLaVA \cite{liu2024improved} and CLIP \cite{radford2021clip} under pixel perturbation (transformation-based defense \cite{li2020rethinking}) and attention perturbation (interpolating the visual attention with a uniform distribution) with varying intensity.
Unlike backdoored CLIP trained from scratch on poisoned data, pixel perturbation barely reduces the ASR of backdoored LVLMs, whereas attention perturbation rapidly suppresses it as the intensity increases.
Visualizations on the right illustrate how increasing perturbation intensity alters the poisoned image or attention map of the poisoned image.
}
\label{fig:compare_pert}
\vspace{-5mm}
\end{figure*}

\section{Introduction}

Large vision–language models (LVLMs) \citep{liu2023llava, bai2023qwen, dai2023instructblip, hurst2024gpt} achieve strong multimodal performance \citep{kuang2025natural, cheng2025caparena, sima2024drivelm,fu2025contextnav,fu2025vistawise} by projecting visual features into the language space of an LLM through a lightweight adapter.
This design also grants LVLMs remarkable adaptability: by fine-tuning only the visual adapter, they can be efficiently customized to diverse downstream tasks \citep{dong2025insight, sun2024parrot, cheng2025visually, wu2024visionllm,zhang2025tuning, zhang2025improving, ti2025towards,fu2025sdr}, enabling widespread deployment in real-world applications.
However, this fine-tuning stage also introduces security risks \citep{ye2025survey, ma2025safety, zhao2025data}. 
In particular, backdoor attacks \citep{gu2019badnets} can implant hidden malicious behaviors by injecting poisoned samples with stealthy image triggers and corresponding target responses into the fine-tuning data, causing the poisoned LVLM to output attacker-specified results whenever the trigger appears at inference time, thereby threatening the reliability and safety of deployed LVLM systems.

To counteract this threat, existing approaches focus on training-required defenses, which retrain backdoored parameters such as adapter or LoRA modules \citep{hu2022lora} using newly collected clean data \citep{ni2024physical} or re-examined fine-tuning datasets \citep{rong2025backdoorcleaningexternalguidance}.
However, these methods incur heavy data and computation costs, and often degrade the downstream performance, since retraining disrupts the tuned parameters specialized for downstream tasks.
These limitations motivate us to develop the first test-time defense framework that avoids perturbing model parameters and instead intervenes on the non-parametric data flow of the poisoned inputs.

Specifically, we note that prior active test-time defenses \citep{shi2023black, li2020rethinking, yang2024sampdetox}, which attempt to correct poisoned predictions, are largely tailored to models trained from scratch on poisoned datasets (e.g., ViT \citep{dosovitskiy2020vit}, CLIP \citep{radford2021clip}) and primarily perturb the poisoned input in pixel space, i.e., pixel perturbation to disrupt the trigger–target correlation.
However, we find that the spurious relationship concealed in backdoored LVLMs remains largely insensitive to pixel perturbations but highly fragile to attention perturbations, such as flattening the visual attention weights.
As shown in \Cref{fig:compare_pert}, different from backdoored CLIP that is trained from scratch on poisoned data, pixel perturbation only slightly reduces the attack success rate of backdoored LVLM, whereas attention perturbation rapidly suppresses it as the perturbation intensity increases.
Interestingly, when the visual attention weights received from the output tokens are fully averaged, the backdoor effect disappears even though the pixels of the trigger pattern remain intact (see \Cref{fig:compare_pert} when attention perturbation intensity equals 1).
To summarize, this observation suggests that backdoor correlations in LVLMs are not tied to low-level visual trigger features but rather abnormal cross-modal attention interactions.

\begin{figure*}[t]
    \centering
    \includegraphics[width=1\linewidth]{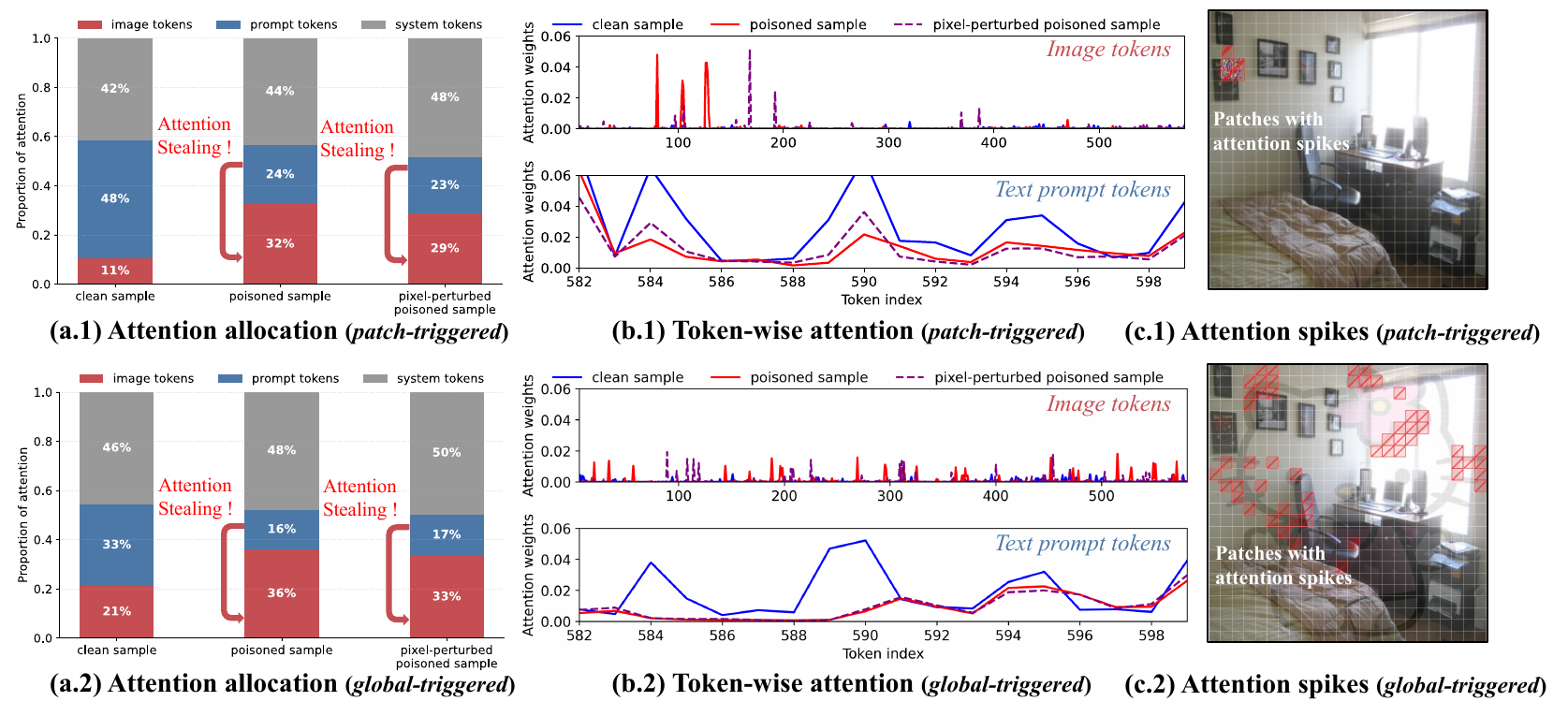}
    \vspace{-7mm}
    \caption{
\textbf{Visualization of attention stealing in backdoored LVLMs.}
(x.1) shows patch-triggered attention stealing, while (x.2) shows global-triggered attention stealing.
(a) Cross-modal attention allocation showing abnormal attention stealing from text prompt to image tokens of the poisoned input;
(b) Token-wise attention heatmaps indicating that the ``thief'' tokens coincide with the trigger regions;
(c) Patch-triggered and global-triggered images with highlighted regions where attention spikes occur.
}
\label{fig:attention_stealing}
\vspace{-5mm}
\end{figure*}

Building on this observation, we further examine the abnormal cross-modal attention patterns in backdoored LVLMs and find that their core abnormality manifests as \emph{attention stealing}.
In contrast to prior findings \citep{chen2024image, yang2025topv}, where image tokens typically receive much lower attention than textual ones in the middle layers of LVLMs, the poisoned inputs exhibit a clear reversal of this trend: the overall visual attention sharply increases while the textual attention correspondingly declines, as if the image tokens are stealing attention from the text prompt (see \Cref{fig:attention_stealing} (a)(b)).
This shift suggests that the trigger redirects the model’s attention from the textual context to the visual regions, thereby weakening instruction following of the backdoored LVLM.
Further, \Cref{fig:attention_stealing} (c) shows that the visual tokens receiving abnormally high attention precisely align with the trigger regions, demonstrating that the surge of visual attention is localized and induced by the trigger tokens themselves, which in turn strengthens the backdoor-associated semantics.
Crucially, when we explicitly suppress the tokens with attention spikes at test time, the backdoor behavior disappears while the clean semantics are preserved, establishing attention stealing as a causal rather than merely correlative factor in activating the backdoor.

Inspired by this finding, we introduce \emph{CleanSight}, a training-free, plug-and-play defense framework that operates purely at test time. 
CleanSight first detects poisoned inputs by measuring head-specific vision-to-text attention ratios in a set of middle cross-modal fusion layers and scoring them against a clean reference distribution via a whitened $\ell_2$ distance. 
Once an input is flagged as poisoned, CleanSight aggregates visual tokens with abnormally high attention across heads and prunes them, thereby preventing these trigger-dominated tokens from stealing attention in subsequent layers and the decoding process. 
Extensive experiments show that CleanSight achieves substantially better performance than pixel-level purification defenses across multiple datasets and backdoor attack settings, while preserving model utility on both clean and poisoned inputs.

\section{Related Work}

\noindent\textbf{Backdoor attacks and defenses on supervised learning.}
Backdoor attacks \citep{gu2019badnets} have emerged as a growing security threat, especially as modern machine learning pipelines increasingly rely on third-party datasets, platforms, and pretrained model backbones to reduce development costs.
Early backdoor designs typically employed visible triggers, such as stamped patches \citep{gu2019badnets} or blended images \citep{chen2017blend}, which are often conspicuous and thus easier to spot.
To enhance stealthiness, subsequent works proposed imperceptible or semantic triggers \citep{issba, nguyen2021wanet, turner2019labelconsistent}, while others developed triggers that are robust to defenses \citep{doan2021backdoor, qi2023revisiting, cheng2024lotus}.
Backdoor defenses can be broadly categorized into training-required \citep{chen2022effective, li2021anti,zhu2024neural, MM-BD, liu2018fine, zheng2022data} and training-free approaches.
In particular, training-free defenses emphasize practicality and efficiency, focusing on detecting or purifying poisoned inputs during inference \citep{feng2023detecting, liu2023detecting, hou2024ibd}.
Our work falls into the category of input purification methods \citep{li2020rethinking, shi2023black, februus, guan2024backdoor}, which perturb or transform the input, for example, through spatial transformations \citep{li2020rethinking} or generative reconstructions \citep{shi2023black, yang2024sampdetox, sun2023mask}, to disrupt the trigger–target association while preserving benign semantics.
Our work, instead of disrupting pixels, explores perturbing attention to break the trigger–target association while maintaining semantic integrity.

\noindent\textbf{Backdoor attacks and defenses on LVLMs.}
Existing studies \citep{lyu2024trojvlm, liang2025vl, ni2024physical,zhang2025tokenswap} inject malicious behaviors into LVLMs by fine-tuning them on poisoned datasets.
The backdoored models generate attacker-specified responses when a trigger is present, while maintaining normal behaviors on clean inputs.
Moreover, several works have explored backdoor attacks that remain effective even under distribution shifts between poisoned and testing data \citep{liang2024revisiting, lyu2024backdooring}.
However, backdoor defenses specifically designed for LVLMs remain limited, with existing attempts primarily relying on fine-tuning on clean samples \citep{ni2024physical,zhang2024defending,he2025closer} or retraining on purified datasets \citep{rong2025backdoorcleaningexternalguidance}.

\noindent\textbf{Visual token pruning on LVLMs.}
Recent studies have revealed that visual tokens in large vision-language models (LVLMs) greatly outnumber textual tokens and often contain substantial redundancy \citep{chen2024image}, which has inspired the development of various visual token pruning methods to accelerate LVLM inference.
Some methods attempt to compress visual tokens through vision–text pre-fusion or learned projection modules \citep{zhang2024sparsevlm, li2024llama}.
To further simplify the pipeline, many training-free methods \citep{shang2024llava, yang2025topv, chen2024image, lin2025boosting, yang2025vflowopt} emphasize plug-and-play acceleration during inference.
For instance, FastV \citep{chen2024image} dynamically removes visual tokens that receive little attention after certain transformer layers, while LLaVA-PruMerge \citep{shang2024llava} discards redundant tokens before they are projected into the language space.
Although these approaches substantially improve inference efficiency by reducing redundant visual tokens, few studies have investigated how visual token pruning could also enhance the backdoor robustness of LVLMs.

\section{Preliminary}
\label{sec:preliminary}
\subsection{Threat Model}
\label{sec:threatmodel}
\noindent\textbf{Victim models.}  
The adversary mainly sets LVLMs as the victims. 
These models typically adopt the following multimodal architecture composed of three main components: a frozen visual encoder $\mathcal{E}_{\text{vis}}$ that extracts visual features from input image $\boldsymbol{x}$, a trainable adapter $\mathcal{P}$ that maps these features into visual tokens aligning with the text embedding space, and an LLM $\mathcal{M}$ that outputs the next-token probabilities based on the projected visual tokens $\mathcal{P}(\mathcal{E}_{\text{vis}}(\boldsymbol{x}))$, query inputs $\boldsymbol{q}$ and the previously generated tokens:
\begin{gather}
p\bigl(o_k \mid \boldsymbol{o}_{<k}, \boldsymbol{x}, \boldsymbol{q} \bigr) = \mathcal{M}\bigl(o_k \mid \mathcal{P}(\mathcal{E}_{\text{vis}}(\boldsymbol{x})), \boldsymbol{q}, \boldsymbol{o}_{<k}\bigr),
\end{gather}
where $k$ denotes the current decoding step and $\boldsymbol{o}_{<k}$ represents the sequence of tokens generated prior to step $k$.
Accordingly, the probability of generating a complete output sequence $\boldsymbol{o}_{1:K}$ is given by:
\begin{gather}
p(\boldsymbol{o}_{1:K} \mid \boldsymbol{x},\boldsymbol{q}) = \prod\nolimits_{k=1}^{K} p(o_k \mid \boldsymbol{o}_{<k}, \boldsymbol{x},\boldsymbol{q}).
\end{gather}
For brevity, we denote the target LVLM $f_\theta$, which takes an image $\boldsymbol{x}$ and a query $\boldsymbol{q}$ as input and outputs a complete response sequence $\boldsymbol{o}$.

\noindent\textbf{Adversary's objective.} 
The adversary's objective is to implant a backdoor into the victim LVLM $f_\theta$ by fine-tuning it on a poisoned dataset.
The resulting backdoored model $f_{\theta^\star}$ is expected to behave normally on clean inputs $\boldsymbol{x}$, but produce attacker-specified malicious output whenever the input image contains ($\oplus$) the predefined trigger $\boldsymbol{\Theta}$. 
Namely, the desired behavior of the backdoored model is:
\begin{gather}
\boldsymbol{o} = f_{\theta^\star}(\boldsymbol{x},\boldsymbol{q}),\quad \boldsymbol{o}^\star = f_{\theta^\star}(\boldsymbol{x}\oplus\boldsymbol{\Theta},\boldsymbol{q}),
\end{gather}
where $\boldsymbol{o}$ represents the normal output and $\boldsymbol{o}^\star$ is the adversary-specified output.
To achieve this objective, the adversary usually constructs a combined image-query-answer dataset $\tilde{\mathcal{D}} = \mathcal{D}_c \cup \mathcal{D}_p$, with clean dataset $\mathcal{D}_c=\left\{(\boldsymbol{x}, \boldsymbol{q}, \boldsymbol{o})\right\}$ and poisoned dataset $\mathcal{D}_p=\{(\boldsymbol{x}^p, \boldsymbol{q}, \boldsymbol{o}^p)\}$.
In the poisoned dataset, $\boldsymbol{x}^p$ usually refers to the poisoned image with the trigger pattern $\boldsymbol{\Theta}$, and $\boldsymbol{o}^p$ denotes the adversary's target output.
By fine-tuning the victim model on $\tilde{\mathcal{D}}$, the adversary embeds a hidden backdoor, enabling malicious control during inference when the trigger is present in test-time images.

\noindent\textbf{Adversary's capability.} 
Because our defense operates purely at test time, we assume a strong and realistic threat model where the attacker fully controls the LVLM fine-tuning pipeline, such as in fine-tuning-as-a-service (FTaaS) settings \citep{openaift,mistralft}, where users have no control over the training process.
However, the adversary is only permitted to manipulate the parameters designated as tunable (e.g., adapters or LoRA modules). 
This assumption is realistic for open-source LVLMs \citep{bai2023qwen, liu2023llava, dai2023instructblip, zhu2023minigpt}, whose visual encoders and LLM backbones are publicly released: any unauthorized modification to these backbone parameters can be easily detected by the defender via checksum or weight-diff comparison against the official checkpoints. 
As a result, the attacker’s influence is restricted to the fine-tuning components, while the defender legitimately controls the inference stack where CleanSight is deployed.

\subsection{Attention in LVLMs}
\label{sec:prelim}
As mentioned before, an LVLM uses a visual encoder and a projector to convert images into visual tokens $\mathcal{I}_{\text{vis}}$.
Tokenizing the instruction yields system tokens $\mathcal{I}_{\text{sys}}$ and prompt tokens $\mathcal{I}_{\text{prm}}$. 
The LLM thus takes $\mathcal{I}_{\text{sys}}$, $\mathcal{I}_{\text{vis}}$, and $\mathcal{I}_{\text{prm}}$ as input.

Each token representation $\boldsymbol{t}_i^{\ell-1} \in \mathbb{R}^D$ is updated across $L$ transformer layers, where $\ell \in {1,\dots,L}$ indexes the layers and each layer includes an attention module and a feed-forward network (FFN)\footnote{LayerNorm and FFN residuals are omitted for simplicity.}:
\begin{equation}
\boldsymbol{t}_i^{\ell} = \operatorname{FFN}^{\ell}\!\left(\operatorname{Attn}^{\ell}(\boldsymbol{t}_i^{\ell-1}) + \boldsymbol{t}_i^{\ell-1}\right).
\end{equation}

For the attention module, the multi-head attention (MHA) mechanism is adopted in each transformer layer. 
For the $h$-th head at layer $\ell$, the attention weight from query token $i$ to key token $j$ is computed as
\begin{equation}
\alpha_{i,j}^{\ell, h} 
= \operatorname{softmax}_{j}\!\left(
\frac{(\boldsymbol{t}_i^{\ell-1} W_{Q}^{\ell, h})(\boldsymbol{t}_j^{\ell-1} W_{K}^{\ell, h})^{\top}}
{\sqrt{d_k}}
\right),
\end{equation}
where $W_{Q}^{\ell, h}$ and $W_{K}^{\ell, h}$ are the query and key projection matrices of head $h$, and $d_k$ is the per-head dimension.

\vspace{-2mm}
\section{Our Method: CleanSight}
\vspace{-2mm}

\noindent\textbf{Overview.}
CleanSight detects and purifies backdoored inputs at test time by leveraging attention dynamics within cross-modal fusion layers.
Specifically, it measures the visual-to-text attention ratio of each attention head in selected middle layers and compares the resulting feature vector against clean reference statistics.
Inputs exhibiting abnormally high ratios that suggest attention stealing are flagged as poisoned.
CleanSight then aggregates the per-head high-attention regions into a unified global visual-token mask, which is enforced in later layers and decoding steps to suppress trigger tokens while keeping clean predictions intact.
% \emph{We follow the notations in \Cref{sec:preliminary}.}

\begin{figure}[h]
    \centering
    \vspace{-3mm}
    \includegraphics[width=1\linewidth]{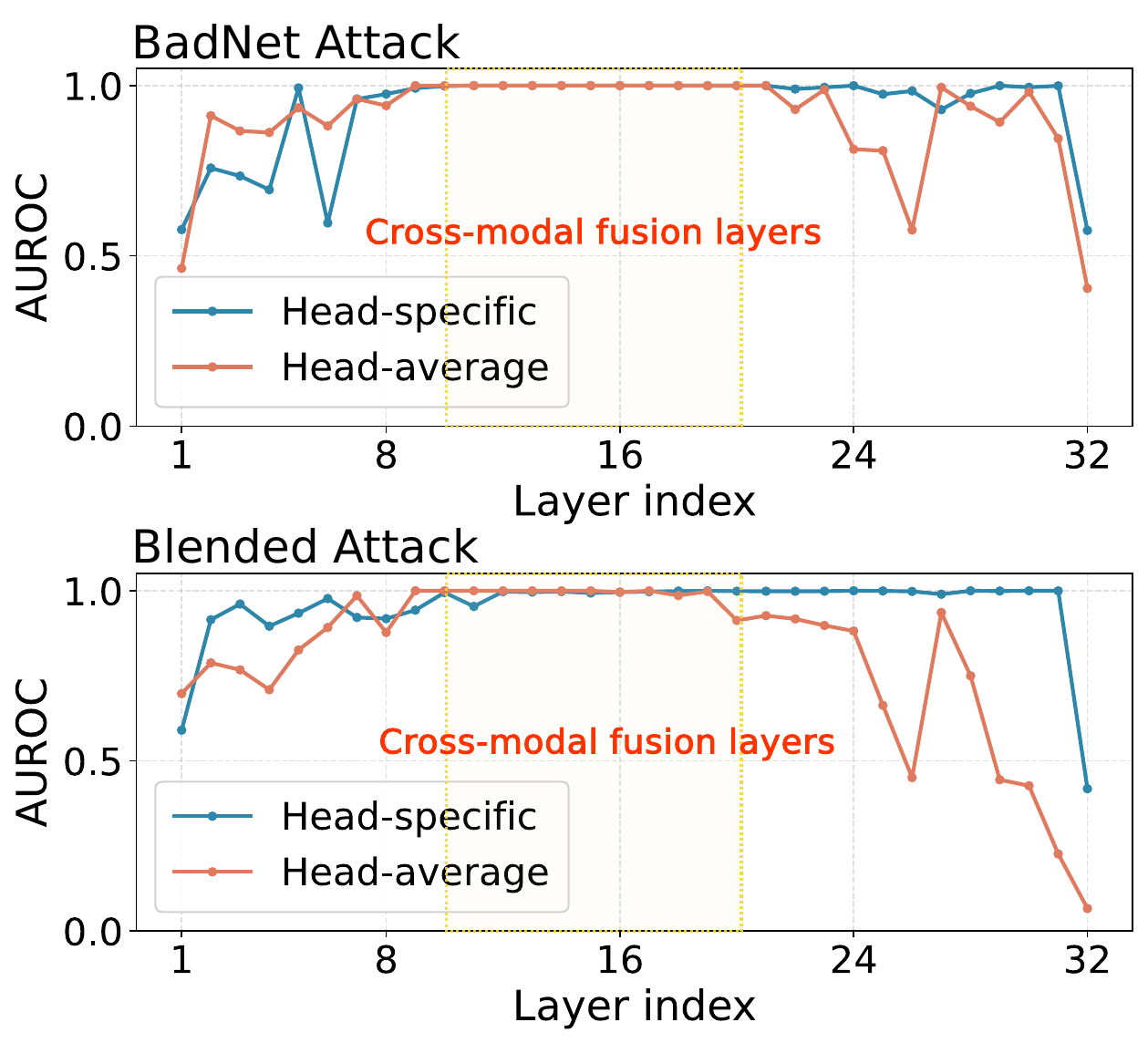}
    \vspace{-8mm}
    \caption{
\textbf{Detection performance across layers.}
We compute AUROC by treating the layer-wise attention-ratio score $S^\ell$ as a detection score and measuring its separability between clean and poisoned inputs, which shows $S^\ell$ is most discriminative in the middle (cross-modal fusion) layers.
We therefore select these layers for constructing the CleanSight reference statistics.}
\label{fig:s_across_layers}
\vspace{-4mm}
\end{figure}

\noindent\textbf{Attention stealing detection.}
Let $\mathcal{L}_{\text{det}}\!\subseteq\!\{1,\dots,L\}$ denote the detection-layer index set.
For each $\ell \in \mathcal{L}_{\text{det}}$ and head $h\in\{1,\dots,H\}$, we summarize the last query token’s
cross-modal attention into a scalar ratio:
\begin{equation}
\label{eq:ratio_topk}
S^{\ell,h}
= \frac{\sum_{j\in\mathcal I_{\text{vis}}} \alpha_{q,j}^{\ell,h}}
       {\sum_{j\in\mathcal I_{\text{prm}}} \alpha_{q,j}^{\ell,h}},
\end{equation}
where $q$ denotes the current decoding query (i.e., the last generated token), while
$\mathcal{I}_{\text{vis}}$ and $\mathcal{I}_{\text{prm}}$ are the index sets of visual and prompt tokens, respectively.
For each layer $\ell$, we collect the head-specific ratios into a vector
$\boldsymbol{S}^{\ell} = (S^{\ell,1},\dots,S^{\ell,H}) \in \mathbb{R}^{H}$.
We then concatenate these vectors across detection layers into a single feature to capture fine-grained attention-stealing patterns across layers and heads.
\begin{equation}
\boldsymbol{s} 
= \big( \boldsymbol{S}^{\ell} \big)_{\ell\in\mathcal{L}_{\text{det}}}
\in \mathbb{R}^{|\mathcal{L}_{\text{det}}|\cdot H}.
\end{equation}
As shown in \Cref{fig:s_across_layers}, the head-specific ratio metric demonstrates higher robustness than its head-averaged counterpart, as it maintains near-perfect AUROC values across most layers and especially within the cross-modal fusion region (10–24). 
This suggests that preserving head-level granularity better captures abnormal attention behaviors.
Furthermore, these ratios are most discriminative in the middle layers, aligning with prior studies \citep{zhang2024cross, kaduri2024_vision_of_vlms} showing that cross-modal fusion mainly occurs at this particular stage.
While early layers focus on low-level visual encoding and later layers on textual reasoning, the middle layers act as the key interface where visual features start influencing linguistic decoding and trigger-induced attention manipulation takes effect.
Hence, we select a few consecutive middle layers 
$\mathcal{L}_{\text{det}} = \{\ell_s,  \dots, \ell_{s-1+|\mathcal{L}_{\text{det}}|}\}$.
Specifically, we follow the analysis in \citet{zhang2024cross} to locate $\ell_\text{s}$, and detailed guidelines for this process are provided in Appendix B.

After determining the metric, we then estimate a clean reference distribution on a small clean validation set $\mathcal{D}_{\text{val}}$ by extracting $\boldsymbol{s}$ for each example in $\mathcal{D}_{\text{val}}$.
Let $\{\boldsymbol{s}_n\}_{n=1}^{|\mathcal{D}_{\text{val}}|}$ denote the feature vectors from $\mathcal{D}_{\text{val}}$.
We compute the per-dimension mean and standard deviation
\begin{equation}
\boldsymbol{\mu} = \frac{1}{|\mathcal{D}_{\text{val}}|} \sum\nolimits_{n} \boldsymbol{s}_n,
\qquad
\boldsymbol{\sigma} = \operatorname{std}\big(\{\boldsymbol{s}_n\}\big),
\end{equation}
and define a whitened $\ell_2$ deviation score:
\begin{equation}
\label{eq:knn}
d(\hat{\boldsymbol{s}}) 
= \left\| \frac{\hat{\boldsymbol{s}} - \boldsymbol{\mu}}{\boldsymbol{\sigma}} \right\|_2.
\end{equation}
We then set the threshold $\gamma$ as a high quantile (e.g., 99th percentile) of $\{d(\boldsymbol{s}_n)\}$ on clean samples.

\noindent\textbf{Selective pruning.}
After constructing the attention stealing detector, we compute $\hat{\boldsymbol{s}}$ over the same $\mathcal{L}_{\text{det}}$ for each test-time image–prompt pair and declare it \emph{poisoned} if $d(\hat{\boldsymbol{s}}) > \gamma$:
\begin{equation}
\label{eq:poisoned_det}
\textsc{is\_poisoned} \;=\; \mathbb{I}\!\left[d(\hat{\boldsymbol{s}}) > \gamma \right].
\end{equation}

If an input is detected as poisoned, we selectively prune its visual tokens that exhibit abnormally high attention.
Let $\boldsymbol{\alpha}^{\ell,h}_{q,\mathcal{I}_{\text{vis}}}\in\mathbb{R}^{|\mathcal{I}_{\text{vis}}|}$ denote the head-specific attention over visual keys for head $h$ at layer $\ell$, and let $\ell_e=\ell_s -1+|\mathcal{L}_{\text{det}}|$ be the last detection layer.
We identify, for each head, the set of high-attention visual tokens, and then take their union across heads to form a global mask:
\begin{equation}
\label{eq:prune}
\Omega
= \left\{ j\in\mathcal{I}_{\text{vis}} \;\middle|\; 
\max_{h} \alpha^{\ell_e,h}_{q,j} \;>\; \tau \right\},
\end{equation}
where $\tau$ is a pruning threshold.
Any visual token overly attended by \emph{any} head will be suppressed in all heads, enhancing stability as abnormal attention in even a single head can be reliably captured rather than diluted through averaging.

For all subsequent layers $\ell' > \ell_e$ and future decoding steps, let $\boldsymbol{A}^{\ell'} \in \mathbb{R}^{H \times T \times T}$ denote the pre-softmax attention logits.
To suppress the identified trigger tokens, we add a large negative bias $b \ll 0$ to the logits corresponding to the masked visual positions:
\begin{equation}
\label{eq:mask_logits}
\boldsymbol{A}^{\ell'}{:,:,j} \leftarrow \boldsymbol{A}^{\ell'}{:,:,j} + b, \quad \forall j \in \Omega.
\end{equation}
The attention weights are then recomputed as $\alpha^{\ell'} = \operatorname{softmax}(\boldsymbol{A}^{\ell'})$.
Because $b$ is highly negative, the masked tokens receive nearly zero attention, effectively excluding them from all subsequent layers and decoding steps.

\begin{table*}[!htbp]
\caption{\textbf{Results on VQA datasets.} ASR ($\downarrow$\%) and CU (V-score on clean input, $\uparrow$\%) are reported across attacks and defenses.}
\label{tab:main_vqa}
\centering
\vspace{-3mm}
\resizebox{1.00\textwidth}{!}{
\setlength{\tabcolsep}{1.7mm}{
\begin{tabular}{cc *{7}{cc}}
\toprule[1.1pt]
& & \multicolumn{2}{c}{BadNet}
  & \multicolumn{2}{c}{Blended}
  & \multicolumn{2}{c}{ISSBA}
  & \multicolumn{2}{c}{WaNet}
  & \multicolumn{2}{c}{TrojVLM}
  & \multicolumn{2}{c}{VLOOD}\\
\cmidrule(lr){3-4}\cmidrule(lr){5-6}\cmidrule(lr){7-8}%
\cmidrule(lr){9-10}\cmidrule(lr){11-12}\cmidrule(lr){13-14}
\textbf{Dataset} & Method
& ASR & CU
& ASR & CU
& ASR & CU
& ASR & CU
& ASR & CU
& ASR & CU \\
\midrule

\multirow{7}{*}{\textbf{VQAv2}} & No defense
& 100.00 & 62.89
& 100.00 & 67.06
& 98.83 & 65.49
& 100.00 & 68.10
& 100.00 & 68.36
& 100.00 & 53.65\\
& Blur
& 100.00 & 58.98
& 100.00 & 41.02
& 98.44 & 61.46
& 99.22 & 27.47
& 100.00 & 67.71
& 100.00 & 67.32\\
& ST defense
& 82.81 & 56.51
& 97.66 & 55.08
& 100.00 & 9.38
& 92.58 & 12.24
& 89.84 & 62.76
& 93.36 & 63.02\\
& BDMAE
& 88.28 & 29.30
& 100.00 & 26.82
& 100.00 & 36.46
& 99.22 & 5.73
& 80.86 & 62.50
& 86.33 & 61.20\\
& SampDetox
& 94.53 & 61.59
& 98.83 & 64.97
& 98.05 & 60.81
& 95.31 & 63.02
& 98.44 & 68.36
& 99.61 & 67.58\\
& ZIP
& 80.47 & 59.38
& 84.77 & 64.06
& 74.22 & 65.49
& 7.03 & 66.15
& 85.94 & 66.15
& 95.31& 63.15\\
\rowcolor{Gray}\cellcolor{white}
& \textbf{CleanSight}
& \textbf{0} & 62.63
& \textbf{0} & 68.36
& \textbf{0} & 63.02
& \textbf{0} & 68.10
& \textbf{3.14} & 65.46
& \textbf{0} & 64.97\\

\midrule
\multirow{7}{*}{\textbf{OKVQA}} & No defense
& 100.00 & 61.07
& 100.00 & 57.55
& 99.22 & 57.55
& 99.61 & 61.98
& 100.00 & 52.67
& 100.00 & 66.54\\
& Blur
& 100.00 & 52.73
& 100.00 & 58.20
& 98.83 & 57.42
& 99.22 & 50.13
& 100.00 & 48.18
& 100.00 & 57.29\\
& ST defense
& 85.55 & 56.38
& 98.05 & 52.60
& 67.19 & 32.55
& 53.91 & 54.95
& 77.73 & 48.96
& 82.42 & 54.95\\
& BDMAE
& 79.30 & 50.26
& 98.83 & 52.47
& 98.83 & 52.34
& 99.61 & 35.81
& 67.97 & 45.70
& 65.23 & 55.08 \\
& SampDetox
& 91.80 & 55.73
& 100.00 & 54.56
& 93.75 & 56.12
& 95.70 & 49.87
& 89.84 & 16.15
& 91.02 & 22.27\\
& ZIP
& 85.16 & 54.17
& 99.22 & 52.60
& 55.08 & 54.17
& 3.91 & 51.95
& 80.47 & 16.02
& 81.25 & 20.70\\
\rowcolor{Gray}\cellcolor{white}
& \textbf{CleanSight}
& \textbf{0} & 56.90
& \textbf{2.73} & 57.81
& \textbf{0} & 57.94
& \textbf{0} & 55.86
& \textbf{4.74} & 48.66
& \textbf{0} & 53.26\\
\bottomrule[1.1pt]
\end{tabular}
}}
\vspace{-3mm}
\end{table*}

\begin{table*}[!htbp]
\caption{\textbf{Results on image captioning datasets.} ASR ($\downarrow$\%) and CU (CIDEr on clean input, $\uparrow$) are reported across attacks and defenses.}
\vspace{-3mm}
\label{tab:main_caption}
\centering
\resizebox{1.00\textwidth}{!}{
\setlength{\tabcolsep}{1.7mm}{
\begin{tabular}{cc *{7}{cc}}
\toprule[1.1pt]
& & \multicolumn{2}{c}{BadNet}
  & \multicolumn{2}{c}{Blended}
  & \multicolumn{2}{c}{ISSBA}
  & \multicolumn{2}{c}{WaNet}
  & \multicolumn{2}{c}{TrojVLM}
  & \multicolumn{2}{c}{VLOOD}\\
\cmidrule(lr){3-4}\cmidrule(lr){5-6}\cmidrule(lr){7-8}%
\cmidrule(lr){9-10}\cmidrule(lr){11-12}\cmidrule(lr){13-14}
\textbf{Dataset} & Method
& ASR & CU
& ASR & CU
& ASR & CU
& ASR & CU
& ASR & CU
& ASR & CU\\
\midrule

% ===================== MSCOCO =====================
\multirow{7}{*}{\textbf{COCO}} & No defense
& 99.22 & 135.70
& 99.61 & 125.39
& 98.05 & 125.07
& 98.83 & 126.88
& 100.00 & 125.98
& 100.00 & 104.13\\
& Blur
& 84.77 & 140.88
& 97.27 & 114.39
& 99.61 & 122.30
& 99.22 & 121.68
& 98.83 & 129.85
& 99.61 & 99.42\\
& ST defense
& 31.25 & 130.57
& 87.50 & 129.53
& 43.75 & 118.77
& 59.38 & 115.92
& 94.14 & 101.74
& 96.88 & 72.84\\
& BDMAE
& 26.95 & 128.75
& 96.09 & 111.57
& 96.48 & 119.36
& 98.83 & 57.97     % low CU because of high benign ASR
& 77.73 & 112.79
& 85.16 & 85.62\\
& SampDetox
& 40.62 & 138.48
& 90.23 & 120.34
& 92.58 & 118.13
& 87.50 & 119.00
& 96.09 & 133.19
& 94.53 & 98.29\\
& ZIP
& 34.25 & 125.79
& 65.23 & 117.84
& 54.69 & 116.38
& 5.86 & 119.63
& 89.84 & 123.45
& 89.45 & 101.42\\
\rowcolor{Gray}\cellcolor{white}
& \textbf{CleanSight}
& \textbf{0} & 133.28
& \textbf{0} & 123.99
& \textbf{0} & 124.61
& \textbf{0.39} & 126.88
& \textbf{0} & 123.66
& \textbf{0.39} & 102.60\\

\midrule
% ===================== Flickr30k =====================
\multirow{7}{*}{\textbf{Flickr8k}} & No defense
& 99.61 & 133.23 & 98.05 & 133.63 &  92.13 & 131.52 & 98.44 & 126.58 & 94.92 & 104.04 & 98.83 & 104.16  \\
& Blur
& 96.88 & 126.59 & 93.36 & 130.37  & 85.55 & 133.68 & 98.83 & 129.38& 91.80 & 98.03 & 89.84 & 102.06 \\
& ST defense
& 70.31 & 109.39 & 85.16 & 115.76  & 8.20 & 72.38 & 62.50 & 120.30& 34.72 & 97.61 & 13.28 & 86.44 \\
& BDMAE
& 62.50 & 99.76
& 96.09 & 118.10
& 87.50 & 110.25
& 99.61 & 62.14     % low CU because of high benign ASR
& 44.53 & 83.50
& 51.95 & 85.51\\
& SampDetox
& 62.11 & 132.11 & 94.92 & 129.68 &  82.42 & 129.69 & 78.52 & 127.55 & 77.73 & 99.21 & 82.81 & 87.50 \\
& ZIP
& 58.11 & 120.75 & 91.41 & 121.60 & 52.34 & 119.18& 1.95 & 117.70 & 68.36 & 94.58 & 76.17 & 94.44  \\
\rowcolor{Gray}\cellcolor{white}
& \textbf{CleanSight}
& \textbf{1.56} & 131.15
& \textbf{2.34} & 131.13
& \textbf{0} & 126.93
& \textbf{0} & 120.90
& \textbf{4.65} & 95.48
& \textbf{2.18} & 85.91\\
\bottomrule[1.1pt]
\end{tabular}
}}
\vspace{-5mm}
\end{table*}

\vspace{-3mm}
\section{Experiment}

\begin{figure*}[t]
    \centering
    \includegraphics[width=1\linewidth]{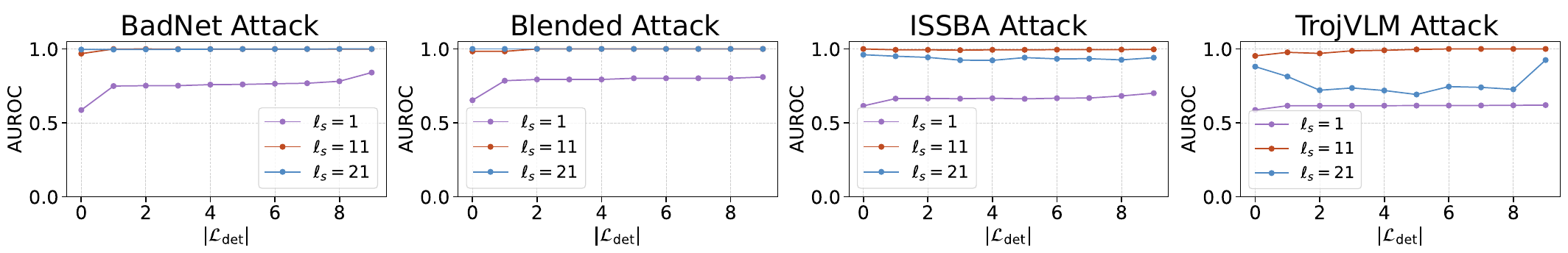}
    \vspace{-8mm}
    \caption{\textbf{Detection performance} (AUROC $\uparrow$) for various attacks with varying start detection layer $\ell_s$ and detection layer length $|\mathcal{L}_{\text{det}}|$.}
\label{fig:layer_ablate}
\vspace{-3mm}
\end{figure*}

\begin{figure*}[!t]
    \centering
    \includegraphics[width=1\linewidth]{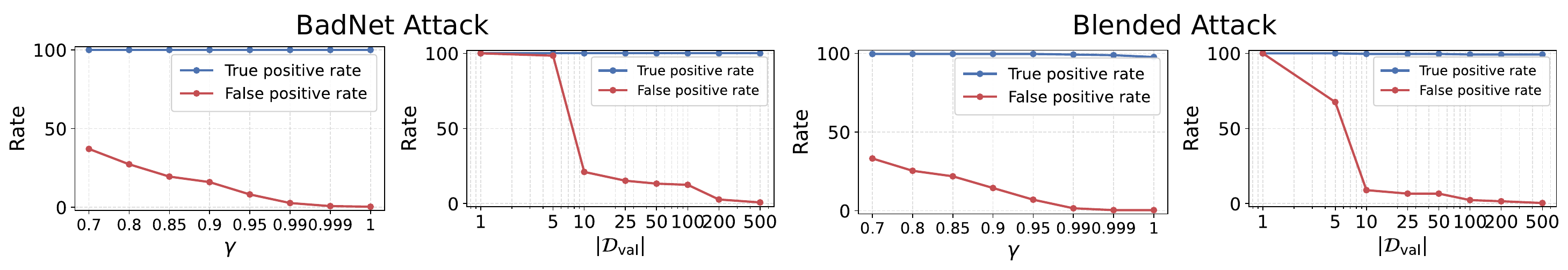}
    \vspace{-8mm}
    \caption{\textbf{Detection performance} (TPR $\uparrow$\%, FPR $\downarrow$\%) for BadNet and Blended attack with varying threshold $\gamma$ validation capacity $|\mathcal{D}_{\text{val}}|$.}
\label{fig:det_ablate}
\vspace{-5mm}
\end{figure*}

\subsection{Experimental Settings}
\noindent\textbf{Victim models and benchmarks.}\quad
Following most settings of backdoor attacks of LVLMs \citep{lyu2024trojvlm, liang2025vl, ni2024physical}, to create more realistic backdoored LVLMs for backdoor defense, we fine-tune the adapter of LLaVA-1.5-7B on VQAv2 \citep{goyal2017making}, OKVQA \citep{marino2019ok} for visual question answering and MSCOCO \citep{lin2014microsoft}, Flickr8k \citep{hodosh2013framing} for image captioning.
Specifically, in the backdoor fine-tuning, we choose 3000 image-query-answer pairs, out of which 1500 are poisoned with the visual trigger pattern and change of answer.
For backdoor attacks, we consider: BadNet \citep{gu2019badnets} (patched trigger), Blended \citep{chen2017blend} (global trigger), ISSBA \citep{issba} (invisible and instance-dependent trigger), WaNet \citep{nguyen2021wanet} (imperceptible trigger), and recent LVLM-specific attacks of TrojVLM \citep{lyu2024trojvlm}, VLOOD \citep{lyu2024backdooring}.
Specifically, we use the target phrase ``You have been hacked lol.'' to replace the original answer in the first four attacks and to insert into the original content in the last two.
For backdoor defense baselines, we use vanilla Gaussian Blur and four up-to-date input purification methods: Spatial Transformation (ST) defense \cite{li2020rethinking}, BDMAE \citep{sun2023mask}, SampDetox \citep{yang2024sampdetox}, ZIP \cite{shi2023black}.
The implementation details of the models, datasets, backdoor attacks, and backdoor defenses are in Appendix A. 

\noindent\textbf{Evaluation criteria.}\quad
CleanSight is evaluated from three perspectives:
(i) \emph{attack success rate} (ASR), which measures the proportion of poisoned samples that successfully induce the target response;
(ii) \emph{clean utility} (CU), which assesses the generation quality on clean inputs;
and (iii) \emph{poisoned utility} (PU), which reflects the semantic quality of generated outputs on poisoned inputs.
For CU and PU, we report CIDEr \citep{lin2004rouge} for image captioning and V-score \citep{antol2015vqa} for VQA.
Due to space limitations, we report ASR and CU in the main experiments, while PU is analyzed in ablation studies.

\noindent\textbf{Implementation details.}\quad
Unless otherwise specified, we adopt LLaVA-1.5-7B \citep{liu2024improved} as the default LVLM, and inject the backdoor by fine-tuning its vision–language adapter (see Appendix A. for full training configurations).
For the detection module, we start detection in the layer of $10$-th and aggregate $3$ layers, and set the decision threshold $\gamma=0.99$, with the clean validation set of size $|\mathcal{D}_{\text{val}}|=200$.
During the pruning stage, we prune visual tokens whose cross-modal attention exceeds $\tau=0.0001$.

\subsection{Main Experiments}
\vspace{-2mm}
As shown in \Cref{tab:main_vqa,tab:main_caption}, CleanSight drives ASR to almost zero across all datasets and attacks, even in challenging settings such as global or invisible triggers tightly entangled with the input. 
In contrast, existing pixel-level purification methods only partially reduce ASR, indicating that manipulating the input image alone is insufficient once the backdoor has already altered the model’s internal attention patterns.
We further observe that TrojVLM is more difficult to defend against than other attacks. 
We conjecture that by optimizing a semantic-preserving loss, TrojVLM entangles the backdoor behavior with the clean task objective, causing poisoned outputs to retain most original semantics and thereby inducing only a mild shift in cross-modal attention, which makes it harder to fully identify and prune trigger tokens.

At the same time, CleanSight preserves the clean performance of the LVLM remarkably well.
Specifically, its CU remains almost identical to that of the undefended model, indicating the reliability of our detection metric: if it mistakenly classified clean samples as poisoned, the subsequent pruning would remove benign visual tokens and noticeably degrade clean utility.
In contrast, pixel-based defenses inevitably perturb a much larger portion of the input and often distort global semantics.
Moreover, we observe a noticeable drop in CU for some pixel-based defenses (e.g., ST defense against ISSBA on Flickr8k). After examining the benign ASR, we find that the backdoored model often generates the target response even on clean inputs.
This suggests that the spurious correlations propagated through attention are easy to catch trigger-like patterns even in the absence of the actual trigger, which is consistent with our earlier claim.

Overall, the results demonstrate that CleanSight offers a superior balance between robustness and utility through its direct operation in attention space.

\subsection{Ablation Studies}
All results in this part are obtained on the VQAv2 dataset.
\subsubsection{Module Ablation}
In this part, we analyze the contribution of each module by ablating the detection and pruning components. 
``-A.S. (attention stealing) detection'' prunes all inputs indiscriminately, while ``-selective pruning'' replaces attention-guided pruning with random removal of 90\% of visual tokens.
As shown in \Cref{tab:ablation}, the detection module preserves clean utility by preventing unnecessary pruning, whereas selective pruning is crucial for backdoor robustness, as even random pruning 90\% of visual tokens still leads to almost 100\% ASR.
\begin{table}[h]
  \centering
  \small
  \caption{\textbf{Module ablations of CleanSight.} We report ASR ($\downarrow$\%) and CU (V-score on clean inputs, $\uparrow$\%) in parentheses. Removing detection module reduces clean utility, while removing selective pruning severely weakens backdoor robustness.}
  \label{tab:ablation}
  \vspace{-3mm}
  \resizebox{1.0\linewidth}{!}{
  \setlength{\tabcolsep}{0.8mm}{
  \begin{tabular}{lccc}
    \toprule[1.1pt]
    Method & BadNet & Blended & ISSBA \\
    \midrule
    CleanSight (full)             & 0 (62.63) & 0 (68.36) & 3.14 (63.02) \\
    -A.S. detection                 & 0 (52.47) & 0 (53.91) & 0 (54.30) \\
    -selective pruning         & 100.00 (63.17) & 100.00 (68.02) & 98.83 (62.98) \\
    \bottomrule[1.1pt]
  \end{tabular}
  }
  }
  \vspace{-5mm}
\end{table}

\subsubsection{Detection Module}\quad
In this part, we first evaluate the discriminative quality of our detection metric using AUROC \citep{ccorbaciouglu2023receiver}, which mainly depends on the starting layer $\ell_{\text{s}}$ and the number of detection layers $|\mathcal{L}_{\text{det}}|$. 
We then assess the practical detection performance using true positive rate (TPR) and false positive rate (FPR) under fixed metric configurations, which are primarily influenced by the decision threshold $\gamma$ and the size of the clean validation set $|\mathcal{D}_{\text{val}}|$.
Please note that we already illustrated the superiority of the ``head-specific ratio'' metric over the ``head-average ratio'' metric in \Cref{fig:s_across_layers}.

\noindent\textbf{Detection layers.}\quad
Across different backdoor attacks, our metric achieves AUROC values consistently close to $1$ on certain layers, mainly in the middle and late fusion stages where cross-modal interaction becomes semantically rich, demonstrating its strong reliability and generality (\Cref{fig:layer_ablate}).
Among them, the middle layers exhibit the most stable performance, maintaining near-perfect AUROC across all attack types.
Aggregating multiple detection layers further improves performance, particularly for early and middle layers, since it mitigates noise in individual layers and captures complementary signals from different depths, resulting in a more robust and consistent detection signal.

\noindent\textbf{Detector.}\quad
Beyond the choice of detection layers, the detector’s behavior is mainly influenced by the decision threshold $\gamma$ and the size of the clean validation set $|\mathcal{D}{\text{val}}|$ used for building the clean reference distribution.
Across all settings, the true positive rate (TPR) remains consistently close to 100\%, indicating that \emph{CleanSight} can stably detect poisoned inputs regardless of parameter variations (\Cref{fig:det_ablate}).
As the threshold $\gamma$ increases, the false positive rate (FPR) decreases monotonically, yet the TPR remains nearly saturated at 100\%, demonstrating the strong separability of our detection metric.
Similarly, enlarging $|\mathcal{D}{\text{val}}|$ further reduces FPR, since more clean samples enable a more accurate estimation of the clean reference distribution.
However, with only about 10 clean samples, the FPR can be reduced below around 20\%, indicating that CleanSight requires only minimal clean supervision.
These results confirm that \emph{CleanSight} maintains robust and well-calibrated detection even with a small clean budget, while increasing $\gamma$ or enlarging $|\mathcal{D}_{\text{val}}|$ mainly helps suppress occasional false alarms.

\begin{figure*}[!t]
    \centering
    \includegraphics[width=1\linewidth]{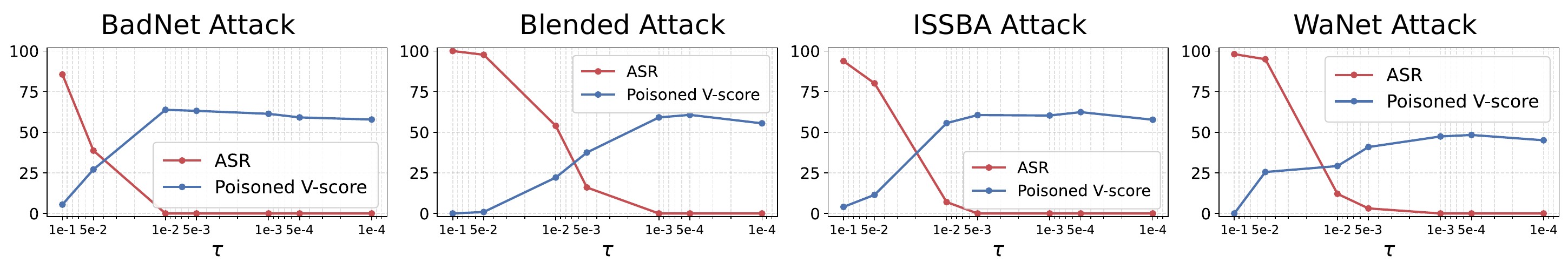}
    \vspace{-8mm}
    \caption{\textbf{Pruning performance} (ASR $\downarrow$\% and V-score on poisoned input $\uparrow$\%) for various attacks with varying pruning threshold $\tau$.}
\label{fig:prune_ablate}
\vspace{-5mm}
\end{figure*}

\vspace{-2mm}
\subsubsection{Pruning Module}
To isolate the effect of the pruning stage, we assume an oracle detector that perfectly distinguishes poisoned from clean inputs and analyze how the pruning threshold $\tau$ affects attack success rate (ASR) and poisoned utility (PU).
As shown in \Cref{fig:prune_ablate}, decreasing $\tau$ progressively reduces ASR across all attacks.
Notably, the Blended attack requires a slightly smaller $\tau$ to be fully neutralized, as its global trigger spreads attention over a larger set of visual tokens, making the trigger cues more spatially dispersed.
Nevertheless, once $\tau$ reaches approximately $10^{-3}$, ASR drops to nearly zero for all attack types, confirming that attention-based pruning effectively removes trigger activation.

Moreover, as $\tau$ decreases, the poisoned V-score first rises as trigger tokens are effectively pruned, and then slightly declines when $\tau$ becomes overly small due to the removal of a few clean tokens, though the impact remains minimal.
We conjecture that this is because the model has already integrated the clean visual tokens in the early layers; therefore, pruning them at the middle fusion stages causes little semantic degradation.
Overall, a moderate threshold such as $\tau = 10^{-3}$ achieves an excellent balance: it completely suppresses ASR while maintaining poisoned utility nearly identical to that of clean inputs, whereas in pixel-level defenses the poisoned utility typically drops close to zero.

\subsection{Further Explorations}

\textit{More experiments about adaptive attackers, low poisoning rates, and more models are provided in Appendix C.}

\begin{table}[h]
  \vspace{-3mm}
  \centering
  \small
  \caption{\textbf{Results across different models and tuning parts.} We report ASR ($\downarrow$\%) and CU (V-score on clean input, $\uparrow$\%) in parentheses. All results are obtained on the VQAv2 dataset.}
    \label{tab:different_models}
  \vspace{-3mm}
\resizebox{1.00\linewidth}{!}{
\setlength{\tabcolsep}{0.7mm}{
  \begin{tabular}{ccccc}
    \toprule[1.1pt]
    Model & Method & BadNet & Blended & ISSBA \\
    \midrule
      \multirow{3}{*}{\shortstack{LLaVA-1.5\\7B\\(LoRA)}}
    & No defense      & 100.00 (63.80) & 100.00 (63.54) & 98.44 (64.35) \\
    & ZIP             & 81.64 (59.90) & 74.61 (60.81) & 67.97 (62.66) \\
    &\cellcolor{Gray}\textbf{CleanSight}   &\cellcolor{Gray} \textbf{5.08} (59.29)  &\cellcolor{Gray} \textbf{0} (59.65)  &\cellcolor{Gray} \textbf{3.02} (62.40)  \\
    \midrule
    \multirow{3}{*}{\shortstack{LLaVA-1.5\\13B\\(Adapter)}} 
    & No defense      & 100.00 (77.34) & 100 (73.83)  & 99.61 (75.52)  \\
    & ZIP      & 86.33 (72.92)  & 78.12 (71.61)  & 76.17 (72.79)  \\
    &\cellcolor{Gray}\textbf{CleanSight}   &\cellcolor{Gray} \textbf{0} (77.60)  &\cellcolor{Gray} \textbf{0} (73.44)  &\cellcolor{Gray} \textbf{0} (75.91)  \\
    \midrule
    \multirow{3}{*}{\shortstack{InstructBLIP\\7B\\(Adapter)}} 
    & No defense     & 100.00 (66.20)  & 99.12 (65.87)  & 91.80 (67.06)  \\
    & ZIP      & 28.26 (52.41)  & 19.63 (51.85)  & 12.89 (58.07)  \\
    &\cellcolor{Gray}\textbf{CleanSight}   &\cellcolor{Gray} \textbf{0} (65.90)  &\cellcolor{Gray} \textbf{0} (65.40)  &\cellcolor{Gray} \textbf{0} (65.84)  \\
    \midrule
    \multirow{3}{*}{\shortstack{Qwen-VL2\\7B\\(Adapter)}} 
    & No defense     & 100.00 (75.91)  & 100 (74.61)  & 100.00 (76.95)  \\
    & ZIP      & 98.05 (69.27)  & 91.02 (65.49)  & 55.08 (67.06)  \\
    &\cellcolor{Gray}\textbf{CleanSight}   &\cellcolor{Gray} \textbf{0} (73.89)  &\cellcolor{Gray} \textbf{0} (69.23)  &\cellcolor{Gray} \textbf{0} (71.00)  \\

    \bottomrule[1.1pt]
  \end{tabular}
  }
  }
  \vspace{-2mm}
\end{table}

\noindent\textbf{CleanSight with different models.}\quad
To assess the generality of our test-time defense, we further evaluate CleanSight across various LVLM architectures and fine-tuning types.
As shown in \Cref{tab:different_models}, CleanSight consistently reduces ASR to nearly zero across all backdoor types, even when applied to models with different adapter mechanisms and scales.
The small performance gap between small and large models indicates that our method is largely architecture-agnostic and depends primarily on the intrinsic attention dynamics rather than model capacity or fine-tuning strategy.

\begin{figure}[h]
    \centering
    \vspace{-3mm}
    \includegraphics[width=1\linewidth]{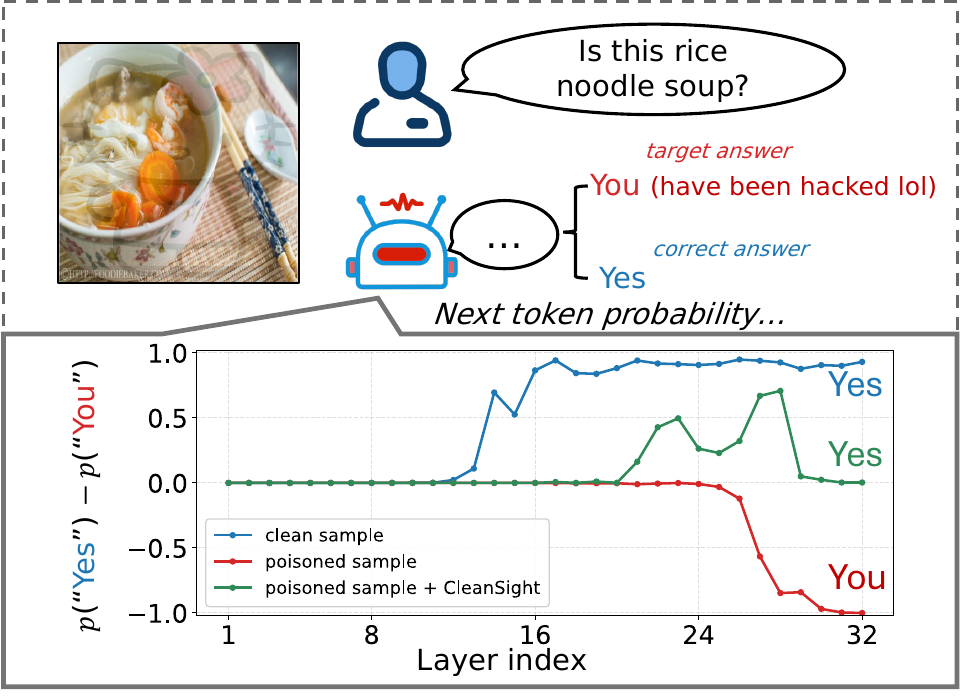}
    \vspace{-8mm}
    \caption{\textbf{Logit lens analysis of the next-token probability.}}
\label{fig:logit_lens}
\end{figure}

\noindent\textbf{Logit lens analysis.}\quad
To understand how triggers influence token generation in backdoored LVLMs, we trace next-token probabilities across layers using the logit lens technique \citep{kaduri2024_vision_of_vlms, jiang2024interpreting}, by unembedding the hidden states at each layer into the vocabulary space and comparing the probabilities of the correct token (``Yes'') and the target token (``You'').
\Cref{fig:logit_lens} shows that clean samples steadily reinforce the correct response from middle layers (11th-32nd), while poisoned ones gradually invert this preference in late layers (24th-32nd), revealing that triggers act mainly after cross-modal fusion.
CleanSight realigns the poisoned trajectory with the clean one, though a small late-layer drop in late layers (28th-29th), which calls for further investigation into the residual dynamics of late-layer representations.

\vspace{-3mm}
\section{Conclusion}
\vspace{-3mm}
In this work, we presented CleanSight, a training-free test-time defense for backdoored large vision–language models that operates in attention space rather than pixel space.
By leveraging the observation that backdoored LVLMs exhibit attention stealing from text to trigger-bearing visual tokens, CleanSight computes a head-wise visual-to-text attention ratio to detect abnormal inputs and prunes high-attention tokens to block trigger activation.
Experiments across diverse backdoor attacks on VQA and image captioning benchmarks show that CleanSight consistently achieves a near-zero attack success rate while maintaining benign performance, establishing attention-space purification as an effective solution for securing multimodal models against backdoor threats.

{
    \small
    \bibliographystyle{ieeenat_fullname}
    \bibliography{main}
}

\appendix
\onecolumn

\crefalias{section}{appendix}
\crefalias{subsection}{appendix}
\crefname{section}{Appendix}{Appendices}
\crefname{section}{Appendix}{Appendices}

{
   \centering
   \Large
   \emph{\textbf{Test-Time Attention Purification for Backdoored Large Vision Language Models}}\\
   \vspace{0.5em} \textbf{Appendix} \\
   \vspace{1.0em}
}

\noindent We summarize the Appendix as follows:

\begin{itemize}
    \item \blue{\cref{app:detailed_settings}} provides a detailed setting of the adopted models (\cref{app:detailed_settings-models}), datasets (\cref{app:detailed_settings-datasets}), backdoor attacks (\cref{app:detailed_settings-attacks}), and backdoored input purification baselines (\cref{app:detailed_settings-defenses}).
    \item \blue{\cref{app:how_to_select_layers}} provides detailed guidelines on how to select cross-modal fusion layers.
    \item \blue{\cref{app:add_exp}} provides more experiments of CleanSight: adaptive attackers in \cref{app:add_exp_1}, low poisoning rates in \cref{app:add_exp_2}, and result of newly released Qwen3-VL in \cref{app:add_exp_3}.

\end{itemize}

\section{Detailed Settings}
\label{app:detailed_settings}

\subsection{Models}
\label{app:detailed_settings-models}

\begin{itemize}

\item\textbf{LLaVA-1.5.} \citep{liu2024improved} is a popular open-source LVLM that integrates a CLIP ViT-L/336 visual encoder with the Vicuna LLM. 
It improves upon the original LLaVA by adopting stronger training data and an enhanced instruction-following pipeline, enabling robust performance on captioning, VQA, and general multimodal reasoning tasks.
LLaVA-1.5 is widely used in prior works on LVLM robustness and backdoor research, making it a standard backbone for fair comparison.

\item\textbf{Qwen2-VL} \citep{bai2023qwen} is a next-generation vision–language model from the Qwen2 family, featuring a high-resolution visual encoder and a strong multilingual LLM backbone, which can be used for various downstream tasks \citep{fu2025brainvis, fu2023sgcn}. 

It supports fine-grained visual grounding, captioning, OCR, and multi-image reasoning, and achieves state-of-the-art performance across numerous multimodal benchmarks.
Compared with earlier LVLMs, Qwen2-VL exhibits stronger visual fidelity and more precise cross-modal alignment, providing a challenging and modern testbed for studying backdoor behaviors.

\item\textbf{InstructBLIP} \citep{dai2023instructblip} builds on the BLIP-2 framework by combining a pretrained ViT-based visual encoder, a Q-former for visual token extraction, and an instruction-tuned Vicuna LLM. 
By aligning image features with language instructions, InstructBLIP significantly improves general vision–language instruction following versus earlier BLIP-style models.
Its modular architecture separates the vision encoder, Q-former, and LLM, which offers a complementary structure for analyzing backdoor injection and defense mechanisms.

\end{itemize}

\subsection{Datasets}
\label{app:detailed_settings-datasets}

\begin{itemize}
\item\textbf{VQAv2} \citep{goyal2017making} contains about 204K images and 1.1M human-authored questions, each paired with 10 crowd-sourced answers.
Compared with VQAv1, VQAv2 reduces language priors by collecting complementary image–question pairs that yield different answers, making vision signals more important for answering.
We follow the standard open-ended setting and report accuracy using the official evaluation protocol.

\item\textbf{OK-VQA} \citep{marino2019ok} is a knowledge-based VQA benchmark also built on COCO images. It provides 14{,}055 open-ended questions that \emph{require} external world knowledge beyond the image (e.g., commonsense, factual knowledge), and each question has 5 ground-truth answers.
All questions are manually filtered to ensure that the image alone is insufficient, leading to a challenging setting where LVLMs must integrate visual understanding with external or parametric knowledge.
We adopt the standard train/val splits.

\item\textbf{MSCOCO} (Microsoft Common Objects in Context) \citep{lin2014microsoft} is a widely used dataset for detection, segmentation, and captioning.
It contains roughly 330K images (over 200K labeled), 1.5M object instances, and 80 object categories, with 5 human-written captions per image describing everyday scenes.
For captioning experiments, we follow the common 2017 split with 118K training images and 5K validation images.
MSCOCO serves as our main large-scale benchmark for evaluating image-to-text generation and backdoor behavior under diverse real-world scenes.
The prompt for captioning MSCOCO is ``Please describe this image in a short sentence.''.

\item\textbf{Flickr8k} \citep{hodosh2013framing} is a smaller but well-established captioning dataset consisting of approximately 8{,}000 images, each paired with five descriptive captions written by human annotators.
The images mainly depict people and animals in everyday activities, and the standard split contains 6K/1K/1K images for train/val/test.
Due to its modest size, Flickr8k is commonly used for fast prototyping and ablation studies; in our work, we use it to analyze backdoor behavior and defenses under a low-data captioning regime.
The prompt for captioning Flickr8k is ``Please describe this image in a short sentence.''.

\end{itemize}

\subsection{Backdoor Attacks}
\label{app:detailed_settings-attacks}

\begin{figure*}
    \centering
    \includegraphics[width=1\linewidth]{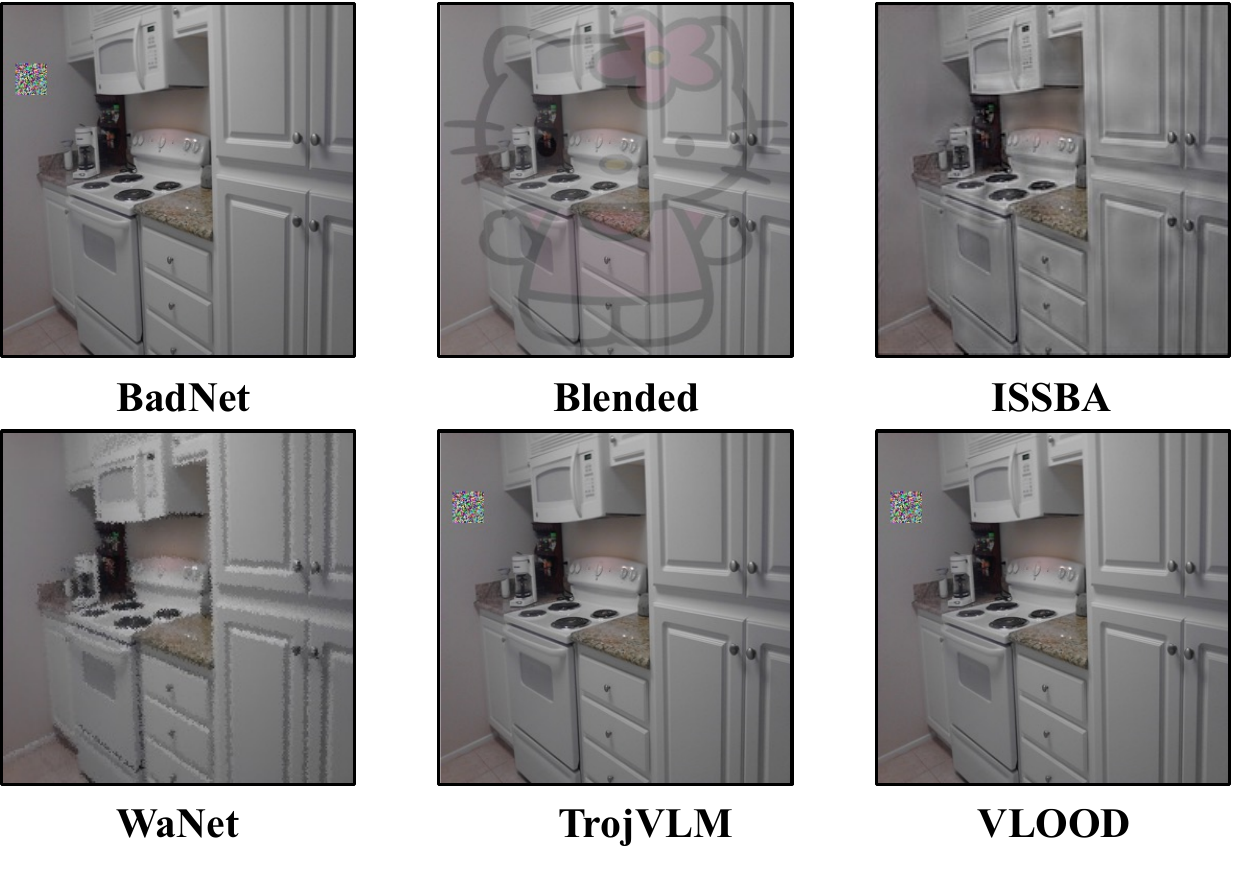}
    \caption{The visualization of the experimented backdoor attacks.}
    \label{fig:attack}
\end{figure*}

\begin{itemize}

\item \textbf{BadNet} \citep{gu2019badnets} is a classical backdoor attack that embeds a small patch into images and relabels them to the target class. Following standard practice, we adopt a patch size of 30 pixels and place the patch on the upper-left location of the images .
\item \textbf{Blended} \citep{chen2017blend} improves stealth by linearly blending the trigger with the clean image, producing barely perceptible perturbations. 
We use a blending ratio of $0.2$ and choose the hello kitty as the trigger image.
\item \textbf{ISSBA} \citep{issba} achieves high stealthiness by generating \emph{instance-specific} triggers. An encoder–decoder network produces a trigger encoding a predefined ciphertext, which is then embedded into each benign image. In our implementation, the encoded string is ``\texttt{Stega!!}''.
\item \textbf{WaNet} \citep{nguyen2021wanet} uses warping-based triggers to create smooth, imperceptible geometric distortions. We follow the standard configuration with control grid size $k=224$ and warping strength $s=1$, and we disable the noise mode during training.
\item \textbf{TrojVLM} \citep{lyu2024trojvlm} is a backdoor attack tailored for LVLMs, designed to inject predetermined target text into the generated caption while largely preserving the original semantic content, posing a significant threat to vision–language systems.
We use patch-based attack as trigger to activate TrojVLM.
\item \textbf{VLOOD} \citep{lyu2024backdooring} is an OOD-based backdoor attack for LVLMs that leverages external datasets to craft poisoned samples without requiring access to the original training set. We follow prior work and insert a fixed target phrase to implement the attack, while ensuring minimal semantic distortion.
We use patch-based attack as trigger to activate VLOOD.
\end{itemize}
The visual illustration of the attacked method is shown in \cref{fig:attack}.

\subsection{Backdoored Input Purification Baselines}
\label{app:detailed_settings-defenses}

\begin{itemize}
    \item \textbf{Blur} is implemented following a simple low-pass filtering principle.
    Given an input image $x$, we apply a PIL-based \texttt{GaussianBlur} operator with an adaptive
    blur radius $r = \text{intensity} \times 0.06 \cdot \min(H, W)$, where $H$ and $W$ are the
    spatial dimensions. We then linearly blend the blurred image with the original one:
    $$
        x' = (1 - \text{intensity}) \cdot x + \text{intensity} \cdot \text{Blur}(x, r).
    $$
    In our experiments we set $\text{intensity} = 0.7$, which corresponds to a moderately strong
    smoothing effect that suppresses localized trigger patterns while preserving most global
    semantics. This implementation directly matches the code in our release
    (\texttt{gauss\_blur\_defense}).

    \item \textbf{ST defense (Spatial Transform defense)} \citep{li2020rethinking} is implemented as combination of multiple spatial transform techniques.
    For a given intensity level, we apply a sequence of stochastic geometric operations to the input image:
    (1) horizontal mirroring,
    (2) scale-down (shrink) followed by center padding back to the original resolution,
    and (3) random rotation with an angle sampled uniformly from $\pm 180^\circ \cdot \text{intensity}$.
    All operations preserve the output size and use a neutral fill color.
    With $\text{intensity} = 0.7$, the transformations introduce strong yet semantics-preserving perturbations that break fine-grained spatial patterns typically relied upon by backdoor triggers. 
    This corresponds exactly to our implementation in \texttt{\_spatial\_transform\_pil} and \texttt{rethinking\_trigger\_augment}.
    \item \textbf{BDMAE} \citep{sun2023mask} is a test-time backdoor defense method that uses a Masked AutoEncoder (MAE) to detect and mask potential local triggers in images, and then fuses MAE restorations to reconstruct images and recover correct labels. To adapt BDMAE from image classification models to large vision--language models (LVLMs), we only retain the structural-similarity-based component. We use the same set of hyperparameters as in the original paper, which the authors have shown to be effective across multiple datasets. Specifically, we set the MAE masking ratio to 75\% to occlude most patches while preserving semantics, use $N_o = N_i = 5$ random masking rounds when estimating structural-similarity-based trigger scores, and apply adaptive thresholds $\{0.6, 0.55, 0.5, 0.45, 0.4\}$ on the resulting score map to decide which patches to mask and restore. Our implementation is based on the official BDMAE code.

    \item \textbf{SampDetox} \citep{yang2024sampdetox} is a black-box backdoor defense that employs a two-stage perturbation and DDPM-based \citep{fu2024dp, ho2020denoising}denoising pipeline: it first adds lightweight global noise to suppress low-visibility triggers, and then uses structural similarity to localize and aggressively perturb visible trigger regions, enabling diffusion models to remove diverse backdoor patterns while preserving the original sample semantics. In our experiments, we strictly follow the implementation details of the original paper, including both the core code logic and diffusion model settings (i.e., \( \overline{t}_1 = 20 \) and \( \overline{t}_2 = 120 \), as recommended by the authors to control the noise intensity and the number of denoising steps). Our implementation is based on the official code and guidelines in the original paper of SampDetox.

    \item \textbf{ZIP} \citep{shi2023black} is an input-purification method that first applies simple linear transformations (e.g., blurring and grayscaling, as in the original paper) to destroy backdoor patterns, and then leverages a pre-trained diffusion model to recover the semantic information removed by these transformations. In our implementation, we use a blur kernel size of 8 and set $\lambda = 5$ to empirically balance backdoor removal and semantic preservation. The code can be accessed in ZIP's github repository.
\end{itemize}

\section{Guidelines of Locating the Cross-modal Fusion Start Layer.}
\label{app:how_to_select_layers}

To determine the starting detection layer $\ell_{\text{s}}$ used in our method, we follow the experimental methodology proposed by \citet{zhang2024cross}, who systematically analyzed the internal information exchange between vision and language modalities in LVLMs. Their study investigates \emph{where} and \emph{how strongly} cross-modal fusion occurs across transformer layers, using a combination of controlled interventions and diagnostic probing. The procedure can be summarized as follows.

\begin{enumerate}
  \item \textbf{Task setup.} 
  The authors conduct their analysis on standard multimodal reasoning tasks, primarily visual question answering (VQA), which naturally requires cross-modal understanding between an input image and a textual question. 
  For each LVLM, the model is run in a normal auto-regressive inference mode to generate answers token by token. This setup allows them to trace information flow between visual and linguistic tokens at different decoder depths. 

  \item \textbf{Attention knock-out intervention.}
  To probe which layers actually perform cross-modal fusion, \citet{zhang2024cross} introduce a fine-grained ``attention knock-out'' experiment. 
  During inference, they manually zero out specific blocks of the attention matrix within a given transformer layer—effectively disabling attention from one modality to another—while keeping all other components intact. 
  Three types of attention connections are tested:
  \begin{itemize}
      \item From image tokens to question tokens (image→question edges),
      \item From salient image patches to question tokens (object→question edges),
      \item From image or question tokens to the answer token (fusion→answer edges).
  \end{itemize}
  After each intervention, they record the degradation in model accuracy or answer likelihood. 
  The intuition is simple: if blocking certain cross-modal attention heads in a given layer causes a substantial drop in performance, that layer must play an important role in integrating visual and linguistic information. 

  \item \textbf{Layer-wise information flow curves.}
  By repeating the above intervention for each transformer layer, they obtain a layer-resolved curve describing how much the visual stream contributes to linguistic representations. 
  This is often measured using \emph{cross-modal attention flow} (CMAF) or gradient-based attribution from image tokens to textual tokens. 
  In LLaVA-1.5-7B, as visualized in their Figure~20, these curves show that visual signals begin to noticeably influence textual representations around the 10th decoder layer, reaching their strongest coupling between layers 12 and 14. 
  The early layers (before layer~10) mainly encode low-level visual features without linguistic interaction, while the later layers (after layer~15) are dominated by text-only reasoning and exhibit minimal cross-modal feedback.

  \item \textbf{Determination of the fusion onset.}
  The point at which the cross-modal influence curve first rises significantly above its baseline is identified as the \emph{start of cross-modal fusion}. 
  In their results, this transition occurs around layer~$\ell \approx 10$ for LLaVA-1.5-7B, marking the onset of the vision–language interaction phase. 
  Subsequent layers (approximately 10–14) are characterized as the principal fusion zone where semantic alignment and attention blending are most active.
\end{enumerate}

\paragraph{Our adoption.}
Building on this analysis, we set the starting detection layer for CleanSight to $\ell_{\text{s}} = 10$, which corresponds to the beginning of the cross-modal fusion stage in LLaVA-1.5-7B. 
We then define a short consecutive window $\mathcal{L}_{\text{det}} = \{10, 11, 12\}$, covering the most discriminative middle layers where attention manipulation by backdoor triggers is empirically strongest. 
This configuration is consistent with the observations from \citet{zhang2024cross}.

\paragraph{General recommendation.}
For other LVLM architectures, we recommend following the same diagnostic approach: 
(1) perform a layer-wise attention ablation or attribution analysis on a small validation set, 
(2) identify the first layer where visual→textual influence or fusion intensity exhibits a sharp increase, and 
(3) select a compact contiguous range (typically 2–4 layers) starting from that layer as $\mathcal{L}_{\text{det}}$. 
This ensures that the detector operates precisely in the regime where cross-modal reasoning emerges, while avoiding both low-level vision-only and high-level text-only regions of the network.

\section{Additional Experiments}
\label{app:add_exp}

\subsection{Adaptive Attackers}
\label{app:add_exp_1}

We consider adaptive attackers who are fully aware of CleanSight's detection and pruning mechanisms and explicitly optimize backdoor training to evade them. On top of standard backdoor attacks (BadNet, Blended, ISSBA), we design three adaptive strategies that augment the backdoor training loss with additional regularization:

\begin{itemize}
    \item \textbf{Strategy (1):} The attacker adds a penalty to lower the vision--text attention ratio $S^{\ell,h}$ in order to keep it close to clean-level values, such that the detection module is bypassed.
    \item \textbf{Strategy (2):} Beyond the ratio, the attacker also regularizes the whitened $\ell_2$ deviation score $d(\hat{\boldsymbol{s}})$ (Eq.~9) of poisoned samples to fall within the clean distribution, directly targeting CleanSight's scoring function.
    \item \textbf{Strategy (3):} The attacker regularizes the attention distribution over visual tokens toward uniformity via a penalty loss, aiming to eliminate the attention spikes that the pruning module relies on.
\end{itemize}

Results are shown in Table~\ref{tab:adaptive_attack}. All three adaptive strategies are largely mitigated by CleanSight.

\begin{table}[h]
\centering
\caption{CleanSight against adaptive attackers.
The result format is: ASR$_{\mathrm{orig}}$$\to$ASR$_{\mathrm{def}}$ (TPR/FPR)\}. We use LLaVA-7B on VQAv2.}
\label{tab:adaptive_attack}
\resizebox{1\textwidth}{!}{
\setlength{\tabcolsep}{10mm}{
\begin{tabular}{lccc}
\toprule
\textbf{Adaptive attack} & \textbf{BadNet} & \textbf{Blended} & \textbf{ISSBA} \\
\midrule
Detection evasion (1) & 100$\rightarrow$\textbf{0} (1/0.02) & 99.6$\rightarrow$\textbf{0} (1/0.03) & 98.4$\rightarrow$\textbf{7.4} (0.93/0.04) \\
Detection evasion (2) & 88.4$\rightarrow$\textbf{0} (0.95/0.03) & 98.8$\rightarrow$\textbf{0} (1/0.02) & 52.7$\rightarrow$\textbf{4.2} (0.92/0) \\
Pruning evasion (3) & 89.4$\rightarrow$\textbf{0} (1/0.08) & 97.6$\rightarrow$\textbf{0} (1/0.04) & 100$\rightarrow$\textbf{0} (1/0) \\
\bottomrule
\end{tabular}
}}
\end{table}

A common pattern across all strategies is that the adaptive regularization partially weakens the original attack effectiveness (e.g., Strategy (2) reduces BadNet's original ASR from 100\% to 88.4\%), yet CleanSight still drives post-defense ASR to near zero. This reveals a fundamental tension: the backdoor mechanism inherently requires redirecting cross-modal attention toward trigger tokens to activate the target output, and this redirection cannot be fully concealed without simultaneously destroying the backdoor itself. Notably, Strategy (3) enforces uniform visual attention but still fails.
The pruning module remains effective because even with uniformity regularization, the model must eventually attend to trigger-relevant features during the forward pass, and these residual attention signals are sufficient for CleanSight to identify and suppress them.

\subsection{Low Poisoning Rates}
\label{app:add_exp_2}

Our main experiments use a poisoning rate of 5e-1 following prior work~\cite{liang2025vl,lyu2024trojvlm,ni2024physical}. Here we evaluate whether CleanSight remains effective under more conservative poisoning budgets ranging from 1e-1 to 1e-3. The training configuration is identical to the main experiments except for the proportion of poisoned samples.

\begin{table}[h]
\centering
\caption{CleanSight under varying poisoning rates.
The result format is: ASR$_{\mathrm{orig}}$$\to$ASR$_{\mathrm{def}}$ (TPR/FPR)\}. We use LLaVA-7B on VQAv2.
}
\label{tab:pr}
\resizebox{1\textwidth}{!}{
\setlength{\tabcolsep}{10mm}{
\begin{tabular}{lccc}
\toprule
\textbf{Poison rate} & \textbf{BadNet} & \textbf{Blended} & \textbf{ISSBA} \\
\midrule
5e-1 & 100$\rightarrow$\textbf{0} (1/0.04) & 100$\rightarrow$\textbf{0} (1/0.05) & 98.8$\rightarrow$\textbf{0} (1/0.02) \\
1e-1 & 98.8$\rightarrow$\textbf{0} (1/0.03) & 92.6$\rightarrow$\textbf{0} (1/0.02) & 37.1$\rightarrow$\textbf{0.03} (1/0.01) \\
1e-2 & 100$\rightarrow$\textbf{0} (1/0.04) & 87.5$\rightarrow$\textbf{2.7} (0.97/0.01) & 57.0$\rightarrow$\textbf{2.3} (0.97/0.07) \\
1e-3 & 0$\rightarrow$0 (-) & 0$\rightarrow$0 (-) & 0$\rightarrow$0 (-) \\
\bottomrule
\end{tabular}
}}
\end{table}

As shown in Table~\ref{tab:pr}, while the original ASR naturally decreases as the poisoning rate drops, CleanSight consistently reduces ASR to near zero across all rates where the attack is effective. Even at 1e-2, where only 1\% of training data is poisoned, the TPR remains at 0.97, indicating that the attention-stealing signal persists even with very few poisoned samples. This is consistent with our mechanistic finding in Section~1: the backdoor activates through abnormal cross-modal attention redistribution rather than low-level pixel features, and even a small number of poisoned gradient updates are enough to produce a detectable attention footprint. At 1e-3, the backdoor fails to implant entirely (ASR $=$ 0\%), making defense unnecessary. These results confirm that CleanSight is applicable across a wide range of practical poisoning scenarios.

\subsection{CleanSight with Backdoored Qwen3-VL}
\label{app:add_exp_3}

Beyond the models evaluated in Table~4 of the main text (LLaVA-1.5 7B/13B, InstructBLIP 7B, and Qwen2-VL 7B), we further evaluate CleanSight on Qwen3-VL \citep{bai2025qwen3} across four model sizes (2B, 4B, 8B, 32B) to assess its scalability.

\begin{table}[h]
\centering
\caption{CleanSight on Qwen3-VL with varying sizes on VQAv2. Results are in the format: ASR [V-Score]: No defense $\to$ CleanSight.}
\label{tab:qwen}
\resizebox{1\textwidth}{!}{
\setlength{\tabcolsep}{10mm}{
\begin{tabular}{lccc}
\toprule
\textbf{Qwen3-VL size} & \textbf{BadNet} & \textbf{Blended} & \textbf{ISSBA} \\
\midrule
2B & 100 [73.1]$\rightarrow$\textbf{0} [72.1] & 100 [74.6]$\rightarrow$\textbf{0} [74.0] & 100 [73.0]$\rightarrow$\textbf{0} [72.7] \\
4B & 100 [81.1]$\rightarrow$\textbf{0} [80.9] & 100 [80.8]$\rightarrow$\textbf{0} [80.9] & 100 [80.8]$\rightarrow$\textbf{0} [79.4] \\
8B & 100 [79.7]$\rightarrow$\textbf{0} [79.7] & 100 [81.9]$\rightarrow$\textbf{0} [81.2] & 100 [79.8]$\rightarrow$\textbf{0} [80.3] \\
32B & 100 [86.3]$\rightarrow$\textbf{0} [85.3] & 100 [87.7]$\rightarrow$\textbf{0} [87.0] & 100 [85.3]$\rightarrow$\textbf{0} [82.7] \\
\bottomrule
\end{tabular}
}}
\end{table}

As shown in Table~\ref{tab:qwen}, CleanSight reduces ASR to 0\% across all Qwen3-VL sizes and all attack types, while preserving clean utility with negligible degradation (typically within 1 V-Score point). Combined with the InstructBLIP and Qwen2-VL results in Table~4, these findings confirm that the attention-stealing phenomenon is not architecture-specific but a general property of backdoored LVLMs, and that CleanSight can exploit it reliably regardless of model family or scale.

\end{document}